\documentclass[11pt,a4paper]{article}
\usepackage[hyperref]{AACL-IJCNLP2020}
\usepackage{times}
\usepackage{url}
\usepackage[T1]{fontenc}
\usepackage{amsmath,amssymb}
\usepackage{latexsym}
\usepackage{bm}
\usepackage{graphicx}
\usepackage{subfigure}
\usepackage{multirow}
\usepackage{makecell}
\usepackage{pifont}
\usepackage{color,soul}
\usepackage{xcolor}
\usepackage{arydshln}

\usepackage{microtype}

\aclfinalcopy 

\title{Named Entity Recognition and Relation Extraction \\ using Enhanced Table Filling by Contextualized Representations}

\author{Youmi Ma \qquad
  Tatsuya Hiraoka \qquad
  Naoaki Okazaki \\
  Tokyo Institute of Technology \\
  \texttt{\{youmi.ma, tatsuya.hiraoka\}@nlp.c.titech.ac.jp} \\
   \texttt{okazaki@c.titech.ac.jp}
 }
\date{}

\begin{document}
\maketitle
\begin{abstract}
In this study, a novel method for extracting named entities and relations from unstructured text based on the table representation is presented. By using contextualized word embeddings, the proposed method computes representations for entity mentions and long-range dependencies without complicated hand-crafted features or neural-network architectures.
We also adapt a tensor dot-product to predict relation labels all at once without resorting to history-based predictions or search strategies. These advances significantly simplify the model and algorithm for the extraction of named entities and relations. Despite its simplicity, the experimental results demonstrate that the proposed method outperforms the state-of-the-art methods on the CoNLL04 and ACE05 English datasets. We also confirm that the proposed method achieves a comparable performance with the state-of-the-art NER models on the ACE05 datasets when multiple sentences are provided for context aggregation.
\end{abstract}

\section{Introduction}
\label{intro}

Named Entity Recognition (NER)~\cite{Nadeau:07,ratinov-roth-2009-design} and Relation Extraction (RE)~\cite{Zelenko:03,zhou-etal-2005-exploring} are two major sub-tasks of Information Extraction (IE).
Recent studies have reported advantages of solving these two tasks jointly in terms of both efficiency and accuracy~\cite{miwa-sasaki-2014-modeling,li-ji-2014-incremental,gupta-etal-2016-table,miwa-bansal-2016-end,zhang-etal-2017-end}. Compared with the pipelined approaches \cite{chan-roth-2011-exploiting}, models that jointly extract named entities (NE) and relations can capture dependencies between entities and relations.

Many existing studies cast joint extraction of NER and RE as a table-filling problem, where entity and relation labels are represented as cells in a single table~\cite{miwa-sasaki-2014-modeling,gupta-etal-2016-table,zhang-etal-2017-end}. As reported by these studies, table-filling is a promising approach for extracting both NE and relations. However, table-filling approaches require feature engineering and search strategy, which is merely a representation of the label space of NER and RE.
Previous work have designed complicated features to encode contexts and long-range dependencies between NE and relations. For example, \citet{miwa-sasaki-2014-modeling} used hand-crafted syntactic features (e.g., the shortest path between two words in the syntactic tree) and \citet{zhang-etal-2017-end} extracted syntactic information using the encoder of a pre-trained syntactic parser.
Authors in \citet{miwa-sasaki-2014-modeling} explore decoding (search) strategies for filling in the table, based on history-based predictions. In addition, they explore six strategies to determine the order of filling for table cells.
History-based predictions are also an obstacle for parallelizing label decoding.

To address the aforementioned issues, we present a novel yet simple method for NER and RE by enhancing \textbf{table}-filling approaches with pre-trained \textbf{BERT}, named \textbf{TablERT}. We utilize BERT initialized with pre-trained weights for representing entity mentions and encoding long-range dependencies among entities to simplify feature engineering. Furthermore, the presented model enhances entity representations with span-based features. To reduce the burden of exploring searching strategies, we utilize a tensor dot-product to fill up cells of relation labels in the table all-at-once (instead of cell-by-cell with beam-search). This modification also simplifies the decoding process and improves decoding parallelism, completing RE with matrix and tensor operations.

This work uses two widely used benchmark datasets, namely, CoNLL04~\cite{roth-yih-2004-linear} and ACE05, both in English, for evaluating the models of both NER and RE. Experimental results demonstrate that the proposed method achieves higher performance than previous state-of-the-art methods, including SpERT~\cite{spert} and DyGIE++~\cite{Wadden2019EntityRA} in addition to the conventional table-filling systems~\cite{miwa-sasaki-2014-modeling,zhang-etal-2017-end}.
We confirm that the tensor dot-product successfully predict relation labels at once without any special search strategy.
Moreover, the proposed method attains comparable performance to the state-of-the-art NER model DyGIE++ when providing multiple sentences as input for context aggregation. The source code is publicly available at \url{https://github.com/YoumiMa/TablERT}.

\section{Proposed Method}

\label{method}

 \begin{figure}[t]
     \centering
     \includegraphics[width=.48\textwidth]{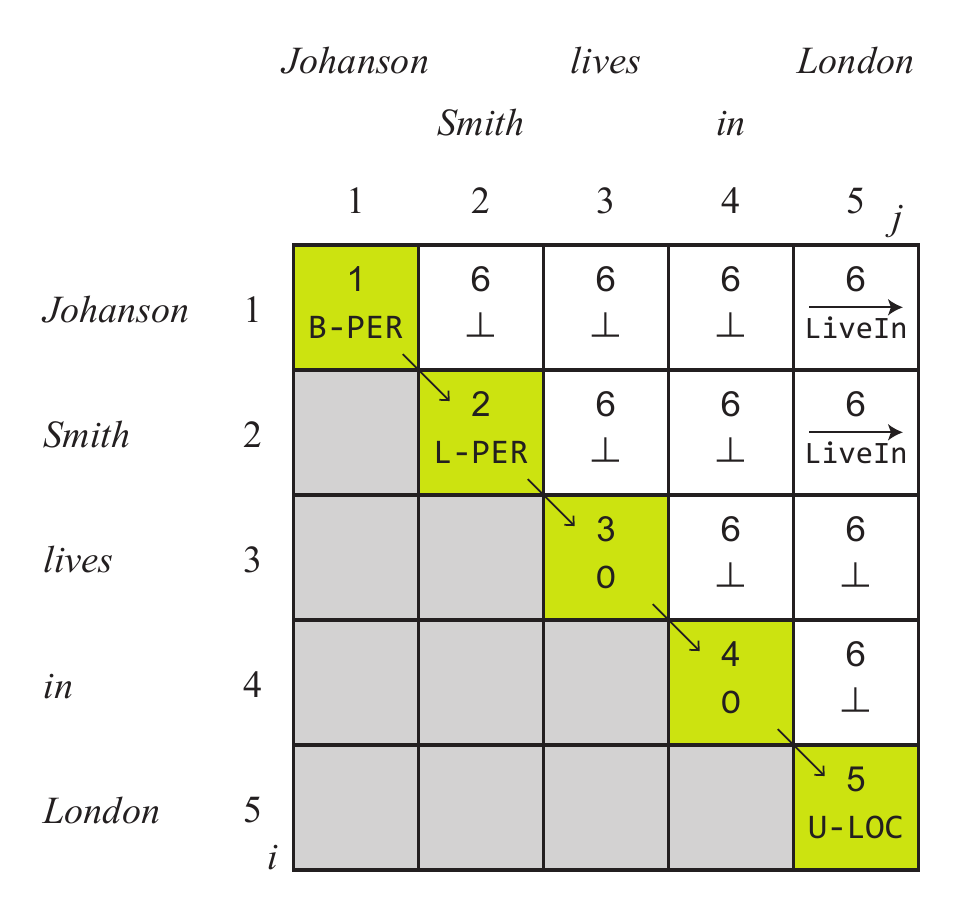}
     \caption{An example of the upper triangular matrix $\bm{Y}$ and table-filling strategy. The numbers in the cells indicate the order of filling cells (decoding). $\perp$ indicates a negative relation label (i.e., there exists no relation among the corresponding words). }
     \label{fig:example}
 \end{figure}

This study aims to extract NE and relation instances. Given a sequence of words $w_1 w_2 \cdots w_n$ ($n$ is the number of words in the input), our goal is to extract relation triples in the form of $\mathrm{(arg0_{type0}, relation, arg1_{type1})}$. Here, $\mathrm{type0}$ represents the NE type of the mention $\mathrm{arg0}$; $\mathrm{arg1}$ and $\mathrm{type1}$ are defined analogously.
We define $\mathcal{E}$ and $\mathcal{R}$ as label sets of named entities and relations, respectively.

The table representation~\cite{miwa-sasaki-2014-modeling} is employed for jointly recognizing NEs and relation instances. Formally, we define an $n \times n$ upper triangular matrix $\bm{Y}$, where a diagonal element $\bm{Y}_{i,i} \in \mathcal{E}$ ($1 \leq i \leq n$) represents an NE label for the word $w_i$, and an off-diagonal element $\bm{Y}_{i,j} \in \mathcal{R}$ ($1 \leq i < j \leq n$) represents a directed relation label between the words $w_i$ and $w_j$. Following \citet{zhang-etal-2017-end}, we hard-code directions into relation labels $\mathcal{R}$ to avoid considering the lower triangular part of the table for RE.
Our model can be seen as a mapping transforming a sequence of words $w_1 w_2 \cdots w_n$ to an upper triangular matrix $\bm{Y}$. We denote an NE label as $y_i = \bm{Y}_{i,i}$ for simplicity.
Figure \ref{fig:example} illustrates an example of a matrix $\bm{Y}$ for the input sentence, ``Johanson Smith lives in London''. Notably, relations are mapped from 1-dimensional word sequences to 2-dimensional matrix $\bm{Y}$ on entity-level. Further, each word inside an entity span is annotated with the corresponding relation label. Take the sentence in Figure \ref{fig:example} as an example, for the NE ``Johanson Smith’’ labeled as \textsc{Person}, relation $\overrightarrow{\textsc{LiveIn}}$ is labeled on both $\bm Y_{1,5}$ and $\bm Y_{2,5}$ corresponding to ``Johanson’’ and ``Smith,’’ respectively. 

\begin{figure}[t]
    \includegraphics[width=.48\textwidth]{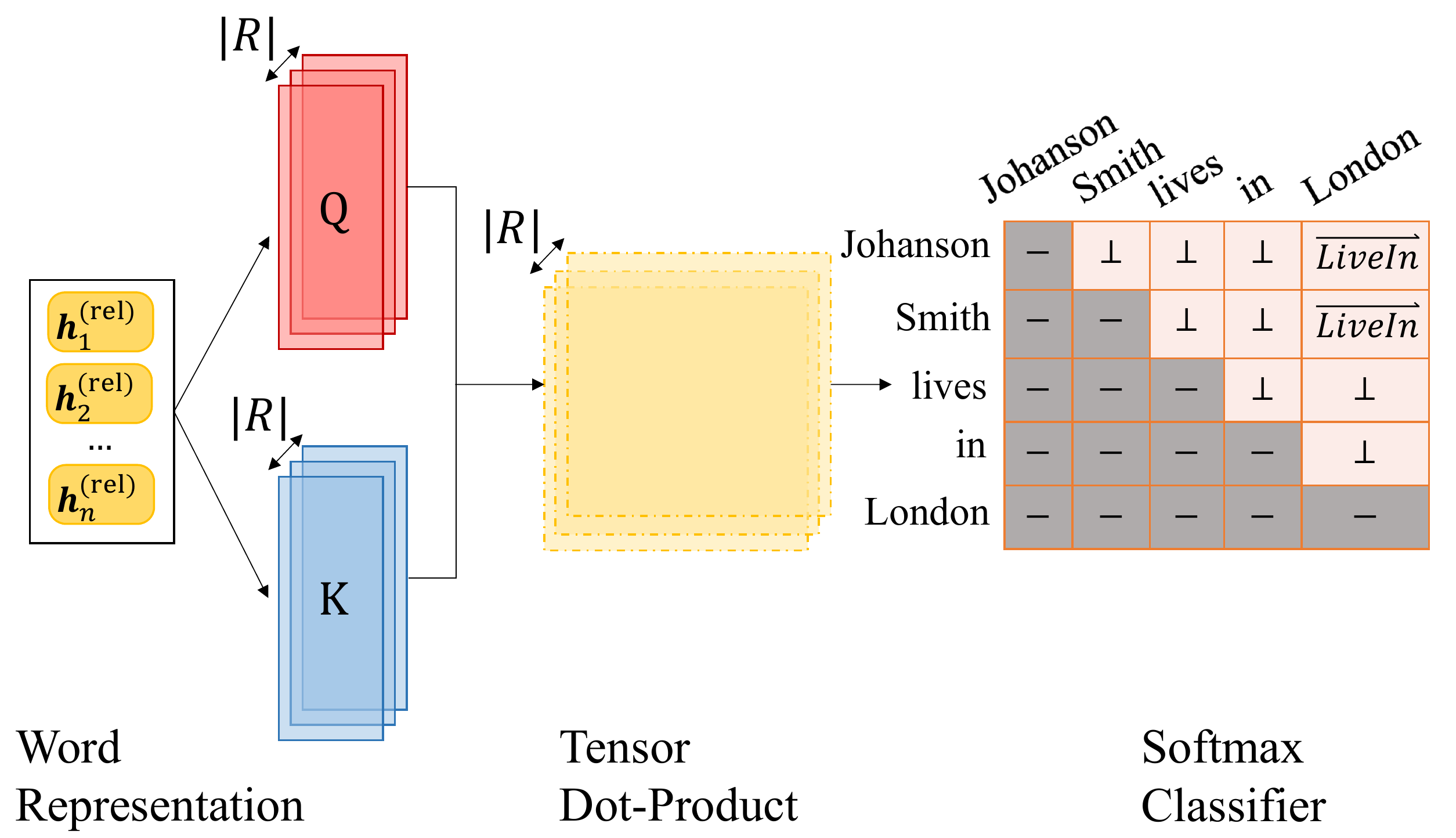}
    \caption{The Relation Extraction model. Relation hidden vectors $\bm{h}_i^{\rm (rel)}$ are computed from the NE module. In the right-most table, $-$ indicates a masked-out cell during both training and predicting.}
    \label{fig:re}
    \end{figure}

This study is based on pre-trained BERT models to leverage contextual information in solving NER and RE. The proposed method stacks layers for NER and RE on the top of a BERT encoder.
As illustrated in Figures \ref{fig:ner} and \ref{fig:re}, our method computes word representations from contextualized embeddings of sub-word tokens obtained by Byte-Pair-Encoding (BPE) (explained in Section \ref{alignment}), and performs NER (Section \ref{ner}) and RE (Section \ref{re}).

\subsection{Word Representations}
\label{alignment}

BERT tokenizer uses WordPiece to split words (e.g., ``Johanson’’) into sub-word tokens (e.g., ``Johan’’ and ``\#\#son’’) with the aid of BPE. This technique is proved to be effective in reducing the vocabulary size and unknown words~\cite{devlin-etal-2019-bert}. Since NEs are annotated at word level, we need its representations at word level during both training and predicting.

In this study, we compute a max-pooling of BERT embeddings of sub-word tokens composing the word as its representation\footnote{We examined the performance on the CoNLL04 development set by using (1) embedding of first sub-word token~\cite{devlin-etal-2019-bert}; (2) mean-pooling of constituent sub-word tokens; (3) mean-pooling of constituent sub-word tokens and \texttt{[CLS]}; (4) max-pooling of constituent sub-word tokens. Among these, max-pooling worked the best.}~\citep{liu-etal-2019-gcdt},
\begin{equation}
    \bm{e}_{w_i} = f(\bm{e}_{t_{i,1}}, \bm{e}_{t_{i,2}}, \cdots, \bm{e}_{t_{i,s}}) .
\end{equation}
Here, we assume the following: the word $w_i$ comprises $s$ sub-word tokens $t_{i,1}, t_{i,2}, \cdots, t_{i,s}$; $\bm{e}_{w_i}$ and $\bm{e}_{t_{i,k}}$ are embeddings for the word $w_i$ and sub-word token $t_{i,k}$, respectively; and $f$ presents a max-pooling operation (henceforth).

\subsection{Named Entity Recognition}
\label{ner}

We use the BILOU (begin, inside, last, outside, unit-length) notation for representing spans of NEs~\cite{ratinov-roth-2009-design}.
We consider NER as a sequential labeling task, where each word $w_i$ in the input is labeled as $y_i$ (a diagonal element in $\bm{Y}$) in the BILOU notation.
In this study, we enhance the existing architecture by using span features at previous timesteps. The use of span features is inspired by \citet{zhang-etal-2017-end}; the authors extracted span representations from bidirectional LSTM cells as features.

    \begin{figure}[t]
     \includegraphics[width=.48\textwidth]{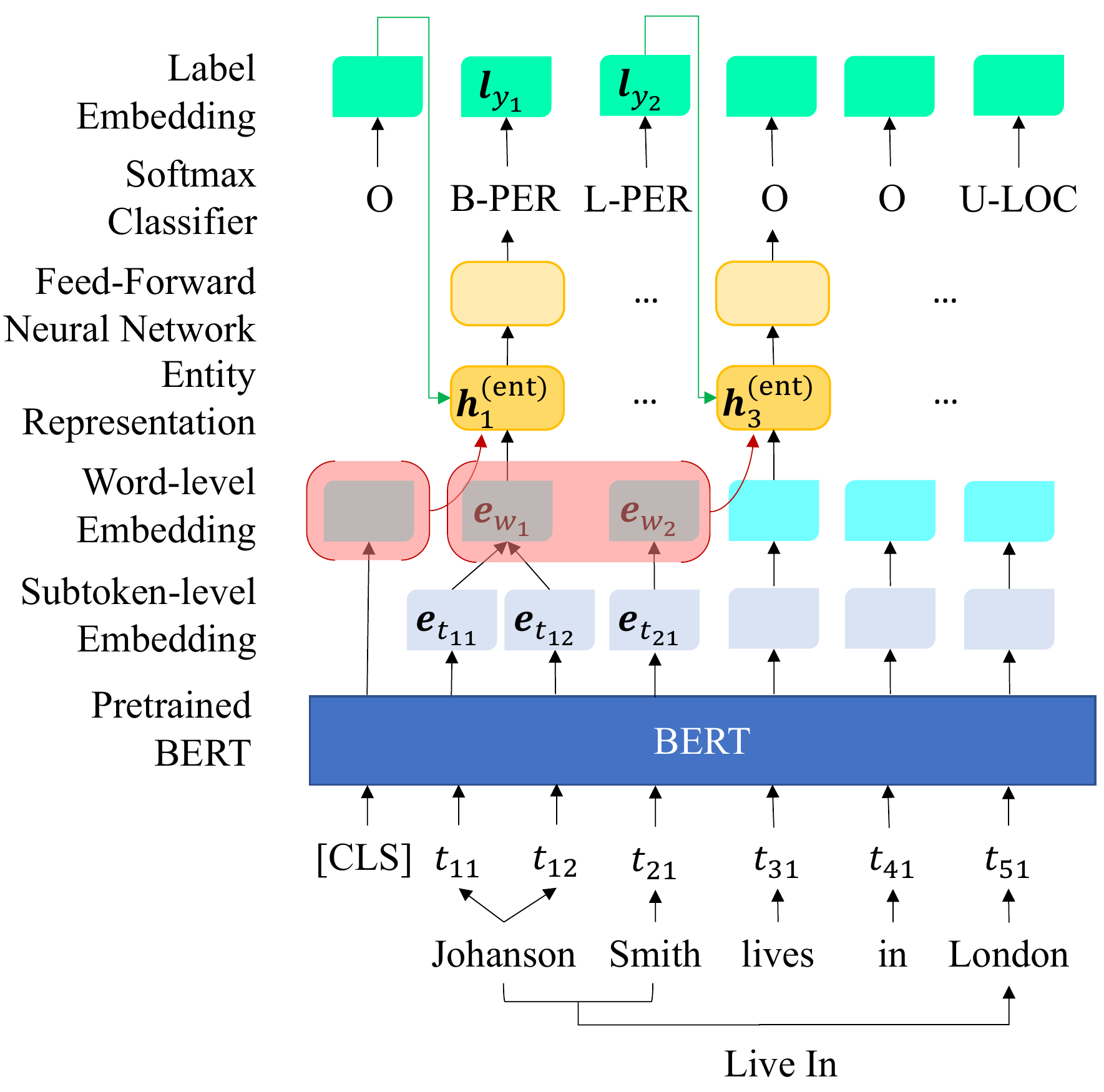}
     \caption{The Named Entity Recognition model. For clarity, we only show the calculation of entity hidden vectors $\bm{h}_1^{\rm (ent)}$ and $\bm{h}_3^{\rm (ent)}$ for the words ``Johanson’’ and ``lives’’ in detail.}
    \label{fig:ner}
    \end{figure}


Specifically, the model predicts an NE label for the word $w_i$ based on three features: (1) a representation $\bm{e}_{w_i}$ of the word $w_i$, (2) embeddings $\bm{l}_{y_{i-1}}$ of the label $y_{i-1}$ at the previous timestep $i-1$, and (3) max-pooling of BERT embeddings of the previous NE span appearing at timesteps $({\rm first}(i-1), \cdots, i-1)$. Here, ${\rm first}(i)$ denotes the timestep where the NE span including word $i$ starts. For example, when processing the sentence shown in Figure \ref{fig:ner}, since the phrase ``Johanson Smith’’ is labeled as an NE, we have ${\rm first}(1)={\rm first}(2)=1$. Similarly, since ``lives’’ is a single non-entity word, we have ${\rm first}(3)=3$. In addition, when the timestep is one ($i=1$), we assume {\tt [CLS]} as the previous word and a unit-length outside (\texttt{O}) as the previous label, i.e., 
$y_0 = \text{\tt O}$ and $w_0 = \text{\tt [CLS]}$. Following \citet{zhang-etal-2017-end}, when the previous word is labeled \texttt{O}, we assume that the previous span is a unit-length span, i.e., ${\rm first}(i-1) = i-1$.

We define the entity representation for predicting the label of current timestep $i$ as the concatenation of three features described above,
\begin{equation}
    \bm{h}_i^{\rm (ent)} = \bm{e}_{w_i} \oplus \bm{l}_{y_{i-1}} \oplus f(\bm{e}_{w_{{\rm first}(i-1)}}, \cdots, \bm{e}_{w_{i-1}}) ,
\label{eq:entity_repr}
\end{equation}
where $\oplus$ stands for a vector concatenation.

We apply a fully connected layer followed by a softmax function $\sigma$ to obtain the probability distribution across all possible NE labels at timestep $i$,
\begin{equation}
    \hat{\bm{y}}_{i} = \sigma(\bm{W}^{\rm (ent)} \bm{h}_i^{\rm (ent)} + \bm{b}^{\rm (ent)}) . \label{equ:ne-prediction}
\end{equation}
Here, $\bm{W}^{\rm (ent)}$ and $\bm{b}^{\rm (ent)}$ represent the matrix and the bias vector for a linear transformation. The vector $\hat{\bm{y}}_{i}$ represents the probability distribution over NE labels; we fill the element $\bm{Y}_{i,i}$ ($=y_i$) with the NE label yielding the highest probability for $\hat{\bm{y}}_{i}$. 
Thus, we perform NER by filling up diagonal elements $\bm{Y}_{i,i}$ from $i=1$ to $n$.

\subsection{Relation Extraction}
\label{re}

We perform RE on top of the BERT encoder and entity spans recognized in Section \ref{ner}.
After the NER model fills up all diagonal elements in $\bm{Y}$, the RE model predicts all off-diagonal elements in $\bm{Y}$.
We adapt a tensor dot-product to score each word pair along with the relation label distribution. The computation is similar to the multi-head self-attention mechanism~\cite{vaswani2017attention}, but our goal is not to compute attention weights over entity representations\footnote{We also tried the deep bi-affine attention mechanism used in \citet{dozat2017deep} and \citet{NguyenV_ECIR2019}. However, we found it sufficient to use the tensor dot-product in the early experiments.}.

Our model utilizes features of entity spans and their NE labels to obtain relation representations for predicting relation labels.
Let ${\rm last}(i)$ denote the timestep where the entity span containing the timestep $i$ ends, analogous to ${\rm first}(i)$ defined in Section \ref{ner}. For instance, for sentence delineated in Figure \ref{fig:ner}, we have ${\rm last}(1) = {\rm last}(2)=2$. The entity-span feature $\bm{z}_i$ (at timestep $i$) is computed using the representations of the constituent words in the entity span,
\begin{equation}
    \bm{z}_i = f(\bm{e}_{w_{{\rm first}(i)}}, \cdots, \bm{e}_{w_{i}}, \cdots, \bm{e}_{w_{{\rm last}(i)}}) .
    \label{eq:pred_z}
\end{equation}
To rephrase, $\bm{z}_i$ is the max-pooling across word representations of the entity span starting at ${\rm first}(i)$ and ending at ${\rm last}(i)$.
Embedding of the NE label $\bm{l}_{y_{i}}$ is then used as the entity-label feature at timestep $i$.
Mathematically, the word representation, i.e., an input to the RE model, is a concatenation of the entity-span and entity-label feature, 

\begin{equation}
    \bm{h}_i^{\rm (rel)} = \bm{z}_i \oplus \bm{l}_{y_i} \label{eq:hrel}.
\end{equation}

For each possible relation $r \in \mathcal{R}$, we apply linear transformations parameterized by two matrices $\bm{W}^{(q)}_{r}, \bm{W}^{(k)}_{r} \in \mathbb{R}^{d_{\rm att} \times d_{\rm rel}}$ and two bias vectors $\bm{b}^{(q)}_r, \bm{b}^{(k)}_r \in \mathbb{R}^{d_{\rm att}}$,
\begin{align}
    & \bm{q}_{i,r} = \bm{W}^{(q)}_{r} \bm{h}_i^{\rm (rel)} + \bm{b}^{(q)}_{r} ,\label{eq:q} \\
    & \bm{k}_{i,r} = \bm{W}^{(k)}_{r} \bm{h}_i^{\rm (rel)} + \bm{b}^{(k)}_{r} ,\label{eq:k} 
\end{align}
where $d_{\rm rel}$ denotes the number of dimensions of $\bm{h}_i^{\rm (rel)}$, and $d_{\rm att}$ represents the number of dimensions after the transformations. $\bm{q}_{i,r} \in \mathbb{R}^{d_{\rm att}}$ and $\bm{k}_{i,r} \in \mathbb{R}^{d_{\rm att}}$ are query and key vectors, respectively, for the relation $r$ at timestep $i$.
As demonstrated, we map the word representation vector $\bm{h}_i^{\rm (rel)}$ into the query and key spaces associated with the relation $r$.

After collecting both query and key vectors for all relations $r \in \mathcal{R}$ at all timesteps $i \in \{1, \dots, n\}$, we obtain two tensors $\mathsf{Q} \in \mathbb{R}^{n \times |\mathcal{R}| \times d_{\rm att}}$ and $\mathsf{K} \in \mathbb{R}^{n \times |\mathcal{R}| \times d_{\rm att}}$.
Slices of the tensors are,
\begin{align}
    & \mathsf{Q}_{i,r,:} = \bm{q}_{i,r} = \bm{W}^{(q)}_{r} \bm{h}_i^{\rm (rel)} + \bm{b}^{(q)}_{r} ,\\
    & \mathsf{K}_{i,r,:} = \bm{k}_{i,r} = \bm{W}^{(k)}_{r} \bm{h}_i^{\rm (rel)} + \bm{b}^{(k)}_{r} .
\end{align}

We compute a probability distribution across all possible relations for every combination of $i$ and $j$ ($1 \leq i < j \leq n)$, which is realized by the dot-product of $\mathsf{Q}$ and $\mathsf{K}$,
\begin{equation}
    \hat{\bm{y}}_{i,j} = \sigma(\mathsf{Q}\mathsf{K}^\top)_{i,j,:} .
    \label{eq:rel_class}
\end{equation}
Here, $(\cdot)_{i,j,:}$ denotes a slice (vector) of the tensor extracting the $(i, j, *)$ elements.
Softmax function $\sigma$ computes the probability distribution across all relation labels $\mathcal{R}$.
We fill in the cell $\bm{Y}_{i,j}$ with the relation label yielding the highest probability for $\hat{\bm{y}}_{i,j}$.
In this way, the RE model predicts relation labels for all pairs of input words at once by computing Equation \ref{eq:rel_class} on the top of NE labels and spans predicted by the NE module.

Meanwhile, we replace Equation \ref{eq:hrel} with \ref{eq:hrel-train} during the training phase,
\begin{equation}
    \bm{z}_i = \bm{e}_{w_{i}} .
\label{eq:hrel-train}
\end{equation}
The rationale behind this treatment is that Equation \ref{eq:hrel} might repeat the same pattern of parameter updates for multiple words within an entity span because of the max-pooling operation.
In contrast, Equation \ref{eq:hrel-train} promotes different patterns of parameter updates for different words in an entity span\footnote{We explore several combinations. For example, using Equation \ref{eq:hrel} for both training and predicting phases, we confirmed that the combination: (Equation \ref{eq:hrel} for prediction and Equation \ref{eq:hrel-train} for training), performed the best.}.

\subsection{Training and Predicting}
\label{training}

The objective of training is to minimize the sum of cross-entropy losses of NER ($\mathcal{L}^{\rm(ent)}$) and RE ($\mathcal{L}^{\rm(rel)}$),
\begin{equation}
    \mathcal{L} = \mathcal{L}^{\rm(ent)} + \mathcal{L}^{\rm(rel)} .
\end{equation}


The proposed NER model uses ground-truth NE labels and spans at training time, similar to \citet{zhang-etal-2017-end}. We perform a greedy search (from left to right) for predicting a label sequence\footnote{The beam search decoding did not depict a definite performance improvement, which is consistent with the report in \citet{miwa-bansal-2016-end}.}. Our RE model receives the predicted NE labels and spans from the NER model and then predicts relation labels based on these predictions. 

Notably, as shown in Figure \ref{fig:example}, if a phrase is labeled as an NE, relations sourcing from or pointing to the phrase will be mapped span-wise to the matrix $\bm Y$. The mapping strategy helps us fully update parameters for different words inside an entity span during relation training. During relation prediction, we ignore inconsistency of label predictions among componential words by max-pooling as in Equation \ref{eq:pred_z}, which results in the same entity-span feature inside the span.

\section{Experiment}
\label{experiments}

\subsection{Datasets}
\label{dataset}


We assessed the performance for NER and RE of our approach TablERT on two widely used datasets: CoNLL04~\cite{roth-yih-2004-linear} and ACE05. In addition, the ability of TablERT to capture cross-sentence dependencies in NER is evaluated on CoNLL03~\cite{tjong-kim-sang-de-meulder-2003-introduction} and ACE05, which include markers for document boundaries.

\paragraph{CoNLL04}  The dataset defines four entity types and five relation types. We report F1 scores for NER and RE adhering to the conventional evaluation scheme. The experiments followed the setup and data split of \citet{gupta-etal-2016-table} and \citet{spert}, which are similar to those of \citet{miwa-sasaki-2014-modeling} and \citet{zhang-etal-2017-end}.
\paragraph{ACE05}  We used the English corpus that encompasses seven coarse-grained entity types and six coarse-grained relation types. We followed the data splits, pre-processing, and task settings of \citet{li-ji-2014-incremental} and \citet{miwa-bansal-2016-end}. For evaluating NER, we regarded an entity mention as correct if its label and the headword of its span were identical to the ground truth. For evaluating RE, we report performance values computed by two different criteria to make them comparable with the previous work: $\text{ACE05}^\diamondsuit$ is indifferent towards incorrect predictions of NE labels, while $\text{ACE05}^\spadesuit$ requires NE labels of relation arguments to be correct.

\paragraph{CoNLL03} The dataset contains four different entity types similar to CoNLL04~\cite{roth-yih-2004-linear}. We used this dataset to measure the performance of NER that considers cross-sentence contexts within a document.

\subsection{Experimental Settings}

Our model is implemented in PyTorch~\cite{pytorch} with HuggingFace Transformer package~\cite{Wolf2019HuggingFacesTS}, utilizing BERT\textsubscript{BASE} (cased)~\cite{devlin-etal-2019-bert} as a pre-trained BERT model.
We ran all experiments on a single GPU of NVIDIA Tesla V100 (16 GiB).
We trained parameters of the NER and RE models as well as those in BERT (fine-tuning) during the training phase, with parameters other than pre-trained BERT initialized with the default initializer.
We used the AdamW algorithm implemented in PyTorch for parameter updates~\cite{loshchilov2018decoupled}.

Hyperparameters were tuned on the held-out development set of CoNLL04. Further, we merged the development and training sets of CoNLL04 for the final training and evaluation, following the procedure of \citet{gupta-etal-2016-table} and \citet{spert}. Major hyperparameters are listed in Appendix A. We report mean values of all evaluation metrics following five runs on each dataset throughout the paper.


\subsection{Main Results}
\label{results}

\begin{table*}[t]
    \centering
    \begin{tabular}{llcccccc}
    \Xhline{3\arrayrulewidth}
         \multicolumn{1}{c}{\multirow{2}{*}{Dataset}}& \multicolumn{1}{c}{\multirow{2}{*}{Model}}  & \multicolumn{3}{c}{Entity} & \multicolumn{3}{c}{Relation}\\
        \cline{3-8}
        & & P & R & F1 & P & R & F1 \\
        \Xhline{2\arrayrulewidth}
        \multirow{5}{*}{CoNLL04} & \citet{miwa-sasaki-2014-modeling} & 81.2 & 80.2 & 80.7 & 76.0 & 50.9 & 61.0 \\
        & \citet{zhang-etal-2017-end}& - & - & 85.6 & - & - & 67.8 \\
        & Multi-turn QA~\cite{li-etal-2019-entity}& 89.0 & 86.6 & 87.8 & 69.2 &  68.2 & 68.9 \\
        & SpERT~\cite{spert}& 88.3 & 89.6 & 88.9 & 73.0& 70.0& 71.5 \\
        \cline{2-8}
        & TablERT (ours) & 89.7 & 90.6 & \textbf{90.2} & 75.0 & 70.3 & \textbf{72.6} \\
        \Xhline{2\arrayrulewidth}
        \multirow{4}{*}{$\text{ACE05}^\diamondsuit$} & \citet{li-ji-2014-incremental} &85.2 & 76.9 & 80.8 & 68.9 & 41.9 & 52.1\\
        &  \citet{dixit-al-onaizan-2019-span}& 85.9 & 86.1 & 86.0 & 68.0 & 58.4 & 62.8 \\
        & DyGIE++~\cite{Wadden2019EntityRA} & - & - & \textbf{88.6} & - & - & 63.4  \\
        \cline{2-8}
        & TablERT (ours) & 87.8 & 88.2 & 88.0 & 70.9 & 61.9 & \textbf{66.1} \\
        \Xhline{2\arrayrulewidth}
        \multirow{5}{*}{$\text{ACE05}^\spadesuit$} & \citet{li-ji-2014-incremental} &85.2 & 76.9 & 80.8 & 65.4 & 39.8& 49.5\\
        & SPTree~\cite{miwa-bansal-2016-end}& 82.9 & 83.9 & 83.4 & 57.2 & 54.0 & 55.6 \\
        & \citet{zhang-etal-2017-end} & - & - & 83.6 & - & - & 57.5\\
        & MRT~\cite{sun-etal-2018-extracting}& 83.9 & 83.2 & 83.6& 64.9 & 55.1 & 59.6\\
        & Multi-turn QA~\cite{li-etal-2019-entity} & 84.7 & 84.9 & 84.8 & 64.8 & 56.2 & 60.2 \\
        \cline{2-8}
        & TablERT (ours) & 87.8 & 88.2 & \textbf{88.0} & 67.0 & 58.5 & \textbf{62.4}\\
        
    \Xhline{3\arrayrulewidth}
    \end{tabular}
    \caption{Micro-averaged precision (P), recall (R), and F1 score (F1) on the test sets of CoNLL04 and ACE05. $\text{ACE05}^\diamondsuit$ regards a relation prediction to be correct when both the relation label and head regions of two arguments are correct. $\text{ACE05}^\spadesuit$ requires that NE labels of arguments are correct in addition to the evaluation criteria of $\text{ACE05}^\diamondsuit$. Notably, NER scores of $\text{ACE05}^\diamondsuit$ and $\text{ACE05}^\spadesuit$ are comparable because the difference in the evaluation criteria affects RE scores only.}
    \label{tab:results}
\end{table*}

\begin{table}[t]
    \centering
    \begin{tabular}{lrr}
    \Xhline{3\arrayrulewidth} 
     \multicolumn{1}{c}{\multirow{2}{*}{Model}}  & \multicolumn{2}{c}{SD}\\
        \cline{2-3}
         & \multicolumn{1}{c}{NER} & \multicolumn{1}{c}{RE} \\
    \Xhline{2\arrayrulewidth} 
         SpERT & 0.378 & 0.857\\
         TablERT (ours) & \textbf{0.187} & \textbf{0.334}\\
    \Xhline{3\arrayrulewidth}
    \end{tabular}
    \caption{Standard derivations (SD) of F1 scores of NER and RE on the CoNLL04 test set predicted by SpERT and this work (with five runs).}
    \label{tab:sd}
\end{table}

Table \ref{tab:results} reports the performance of our method on the datasets, along with in several recent studies on joint NER and RE.
On the CoNLL04 dataset, TablERT achieved comparable or slightly better performance in both NER and RE than SpERT, i.e., an existing state-of-the-art (SOTA) model. Another advantage of TablERT over SpERT is its stability in achieving higher performance. Table \ref{tab:sd} reports the standard derivations (SDs) of F1 scores of NER and RE on the CoNLL04 test set achieved by the two models.

In addition, Table \ref{tab:results} indicates that TablERT outperformed existing work on ACE05.
Regardless of the evaluation criteria for RE ($\text{ACE05}^\diamondsuit$ and $\text{ACE05}^\spadesuit$), F1 scores of TablERT were approximately 1.0 point higher than those of previous SOTA models (DyGIE++ and Multi-turn QA).

TablERT ranked second for NER on ACE05 among previous studies, while DyGIE++ portrayed superior performance on NER. However, their method receives document-level contexts as input that make it incomparable with our model, i.e., trained only with sentence-level contexts. In addition, DyGIE++ utilized coreference information from OntoNotes~\cite{pradhan-etal-2012-conll}.
As we will see in Section \ref{doclevel} with Table \ref{tab:ner}, the proposed method achieved comparable performance to DyGIE++ on NER with document-level context given to the input.

Additionally, a detailed error inspection of both NER and RE is shown in Appendix B. The error inspection aided us in categorizing two significant error types of RE sources, namely, a lack of external knowledge and global constraints.

\subsection{Ablation Test}

\begin{table}[t]
    \centering 
    \small
    \begin{tabular}{lcccccc}
    \Xhline{3\arrayrulewidth} 
    \multirow{2}{*}{Model}  & \multicolumn{3}{c}{Entity} & \multicolumn{3}{c}{Relation}\\
        \cline{2-7}
        & P & R & F1 & P & R & F1\\
        \hline
        Full & \textbf{89.7} & 90.5 & \textbf{90.1} & \textbf{74.7} & \textbf{70.8} & \textbf{72.7} \\
        - Label & \textbf{89.7} & 90.5 & \textbf{90.1} & 74.1 & 70.6  & 72.3 \\
        - Span &   89.5 & \textbf{90.6} & 90.0 & 74.3 & 69.6 & 71.9\\
        - Both &  89.4 & 90.2 &89.8 & 73.5 & 69.7  & 71.3\\
    \Xhline{3\arrayrulewidth}
    \end{tabular}
    \caption{Ablation test of features used in this study on the CoNLL04 test set. ``- Label'' presents the performance when removing the label embeddings at previous timesteps from the model. ``- Span'' removes the features of the previous span. ``- Both'' presents the results when removing both of them.}
    \label{ablation}
\end{table}

To better understand the significance of our proposed method, we conducted ablation tests. New features were gradually removed from the model, and the consequent performance drops were measured.
Specifically, we ablated the features of the label embedding $\bm{l}_{y_{i-1}}$ and the previous span $f(\bm{e}_{w_{{\rm first}(i-1)}}, \cdots, \bm{e}_{w_{i-1}})$ from Equation \ref{eq:entity_repr}.
As can be seen from Table \ref{ablation}, the removal of the features of previous spans from the model had the highest negative impact, whereas the removal of label embeddings at previous time steps had a relatively insignificant negative impact. This result validates our assumption that span-level features are beneficial for representing entities in both NER and RE.

\subsection{Prediction Order}

\begin{table}[t]
    \centering
    \small
    \begin{tabular}{lcccccc}
    \Xhline{3\arrayrulewidth} 
     \multicolumn{1}{c}{\multirow{2}{*}{Order}}  & \multicolumn{3}{c}{Entity}& \multicolumn{3}{c}{Relation}\\
        \cline{2-7}
         & P & R & F1 & P & R & F1\\
    \Xhline{2\arrayrulewidth} 
         Once & \textbf{89.7} & \textbf{90.6} & \textbf{90.2} & \textbf{75.0} & \textbf{70.3} & \textbf{72.6} \\
         Seq & 89.6 & \textbf{90.6} & 90.1 & 74.7 & 70.2 & 72.4\\
    \Xhline{3\arrayrulewidth}
    \end{tabular}
    \caption{Performance of the proposed method on the CoNLL04 test set with history-based predictions. ``Once’’ stands for the proposed method that predicts all relation labels at once. ``Seq’’ decides relation labels sequentially so that it can incorporate prediction results of cells to the left of and below a target.}
    \label{tab:neighbor}
\end{table}

Existing methods for jointly extracting entities and relations with the table-filling approach rely on history-based predictions, i.e., fill up the lower (or upper) triangular part of a table cell-by-cell in a pre-defined order \cite{miwa-sasaki-2014-modeling,gupta-etal-2016-table,zhang-etal-2017-end}.
These methods assume that earlier decisions help later decisions, which may involve long-range dependencies.
In contrast, our model is free from prediction history for relation labels; it focuses on predicting them at once using a tensor dot-product.

A natural question is whether history-based predictions are useful for the proposed method or not?
To find the answer, we designed a variant of the RE model that utilized predicted results of cells to the left of and below a target cell.
More specifically, we modified Equation \ref{eq:rel_class} to make use of the embeddings of relation labels at $(i,j-1)$ and $(i+1,j)$ when predicting a relation label for the element $(i,j)$, and scheduled predictions in ascending order of distance from the diagonal elements and from left-top to right-bottom.


However, a significant improvement in the performance of the model is not observed even after several fine-tuning efforts. Experimental results on the CoNLL04 test set are shown in Table \ref{tab:neighbor}. It is difficult to identify the reason for the experimental results, but the RE model might utilize long-range dependencies from the BERT encoder to make decisions.
In addition, the history-based prediction increases the number of parameters and complexity of label predictions, by introducing extra parameters for relation embeddings and additional classifiers. It is potentially beneficial to try several prediction orders as in \citet{miwa-sasaki-2014-modeling}, but the experimental results suggest that predicting relation labels at once is sufficient for the proposed method.

\subsection{Multi-Sentence NER}
\label{doclevel}


\begin{table*}[t]
    \centering
    \begin{tabular}{lllrrr}
    \Xhline{3\arrayrulewidth} 
    \multicolumn{1}{c}{\multirow{2}{*}{Dataset}} & \multicolumn{1}{c}{\multirow{2}{*}{Model}}  & \multicolumn{1}{c}{\multirow{2}{*}{Sentence}} & \multicolumn{3}{c}{Entity} \\
        \cline{4-6}
        & & & \multicolumn{1}{c}{P} & \multicolumn{1}{c}{R} & \multicolumn{1}{c}{F1} \\
    \Xhline{2\arrayrulewidth} 
        \multirow{5}{*}{CoNLL03} & BERT (reported in \citet{devlin-etal-2019-bert}) & Multi & - & - & 92.4\\
        & BERT (our replication) & Single & 89.5 & 89.9 &  89.7\\
        & BERT (our replication) & Multi & 91.3 & 92.7 & 92.0 \\
        \cline{2-6}
        & TablERT (ours) & Single & 90.3 & 90.5 & 90.4\\
        & TablERT (ours) & Multi & \textbf{92.0} & \textbf{92.9} & \textbf{92.5} \\
        \Xhline{2\arrayrulewidth} 
        \multirow{3}{*}{ACE05} & DyGIE++~\cite{Wadden2019EntityRA} & Multi & - & - & 88.6\\
        \cline{2-6}
        & TablERT (ours) & Single & 87.2 & 88.1 &  87.6\\
        & TablERT (ours) & Multi & \textbf{88.8} & \textbf{88.6} & \textbf{88.7} \\
    \Xhline{3\arrayrulewidth}
    \end{tabular}
    \caption{Results of NER on the CoNLL03 and ACE05 test sets. Values of \citet{devlin-etal-2019-bert} and \citet{Wadden2019EntityRA} are reported scores. We included a baseline BERT, following the original study settings, which involves each word to be represented by its first sub-word token.}
    \label{tab:ner}
\end{table*}

As described in Section \ref{results}, the proposed method could not outperform DyGIE++~\cite{Wadden2019EntityRA} on ACE05 NER, because of the unavailability of cross-sentence information such as coreferences. In this subsection, we describe how we eliminated the performance gap by merely providing multiple sentences into the model without modifying the architecture. Specifically, we split a document into segments of multiple sentences such that each segment was not longer than 256 sub-word tokens. This length restriction is introduced due to GPU memory capacity limitations. Assuming that each segment was a sequence of sub-word tokens from multiple sentences, we fed each segment to the model with multiple sentences separated by a special token {\tt [SEP]}, similar to \citet{devlin-etal-2019-bert}.


Table \ref{tab:ner} shows the performance of NER models on the CoNLL03 and ACE05 with and without multi-sentence inputs. To better investigate the effectiveness of our model, we use ``BERT (our replication)’’ as the baseline. Our models then equip the BERT encoder with extra modules, as described in Section \ref{ner}.
We observe that multi-sentence inputs boosted the performance on both datasets, making our model outperform other models, including DyGIE++ and BERT~ \cite{devlin-etal-2019-bert}.
By comparing the prediction results, we conclude that multi-sentence input improved predictions for multiple occurrences of the same entity, gathering contexts in different occurrences. Specific shreds of evidence with examples are shown in Appendix C.

Although we cannot apply this technique directly to RE (because the table size is $\mathcal{O}(n^2)$), we tested our RE model in a pipelined fashion by using predictions of NER with multi-sentence input on the ACE05 test set. However, we did not see a performance boost on RE. By analyzing predicted RE instances, we discovered that multi-sentence NER increased the overall performance by capturing coreferences, but it also failed to correctly label some of the NEs essential for RE. The observation emphasizes the importance of improving the design of the RE model and a better approach to combine sentence-level and document-level context during RE.

\section{Related Work}
\label{related work}


Early studies formulated the task of jointly extracting entities and relations as a structured prediction with the global features and search algorithms.
\citet{li-ji-2014-incremental} presented an incremental algorithm for joint NER and RE with global features and inexact (beam-search) decoding.
\citet{miwa-sasaki-2014-modeling} proposed a table representation for entities and relations. Further, it investigated hand-crafted features and complex search heuristics on the table.
\citet{gupta-etal-2016-table} enhanced the table-filling approach by adapting recurrent neural networks (RNNs) to fill cells of a table in a pre-defined sequential order.
\citet{miwa-bansal-2016-end} explored a shared representation for entities and relations by stacking bidirectional tree-structured and sequential LSTM-RNNs.
\citet{zhang-etal-2017-end} integrated a global optimization technique and syntax-aware word representations.
These studies heavily relied on feature engineering (as hand-crafted features or specialized models of deep neural networks) and search/optimization strategies.

Recently, several researchers explored the deep contextualized word representations for the sequential labeling problem.
\citet{liu-etal-2019-gcdt} proposed a deep transition architecture enhanced with the global context and reported improvements on NER and chunking tasks by using contextualized word embeddings.
\citet{strakova-etal-2019-neural} also demonstrated the effectiveness of the contextualized representations on the architectures for nested named entity recognition, where NE may overlap with multiple labels assigned.

More recently, span-enumeration methods have been a popular approach for jointly extracting entities and relations~\cite{luan-etal-2019-general,Wadden2019EntityRA,spert}.
In general, span enumeration methods consider possible entity spans for an input sentence with some criteria (e.g., the maximum number of words), choose likely spans using features extracted from the span candidates.
\citet{luan-etal-2019-general} proposed a general framework of information extraction called DyGIE that can incorporate global information on a dynamic span graph. \citet{Wadden2019EntityRA} further expanded the model to DyGIE++. The method receives multiple sentences from the same document as input and enumerates candidate spans for relation, coreference, and event identification. To update span representations of entities, they carefully designed strategies for dynamic graph construction and span refinement.
\citet{spert} also proposed an end-to-end RE model for extracting both entities and relations called SpERT. The method bases on pre-trained BERT models and enumerates candidates of entity spans. Using a negative sampling strategy for both NER and RE, the method classifies entity and relation candidates into positive and negative.

We aimed at solving the drawbacks of the table-filling approach, e.g., complicated feature engineering and decoding algorithm. In our approach, feature engineering for NER was removed by using contextualized word representations and span-based features. The proposed method utilized a tensor dot-product for filling in off-diagonal cells at once without using history-based predictions.
Although the proposed architecture was different from the span-enumeration based approaches (DyGIE++ and SpERT), the experimental results demonstrated competitive or better performance than the span-enumeration based approaches.

\section{Conclusion}
\label{conclusion}

This paper presented TablERT, a novel method for extracting NE and relations based on the table representation, making use of contextualized word embeddings for representing entity mentions. We applied tensor dot-product for predicting all the relation labels at once. The experimental results on the CoNLL04 and ACE05 dataset demonstrated that the proposed method outperformed not only the existing table-filling methods but also the SOTA methods based on pre-trained BERT models. We also confirmed that the method achieved comparable performance to the SOTA NER models on the ACE05 when multiple sentences were fed to the model.

In the future, we plan to explore an approach for incorporating global constraints in the RE model, which currently predicts all relation labels independently. 


\bibliography{aacl-ijcnlp2020}

\begin{thebibliography}{28}
\expandafter\ifx\csname natexlab\endcsname\relax\def\natexlab#1{#1}\fi

\bibitem[{Chan and Roth(2011)}]{chan-roth-2011-exploiting}
Yee~Seng Chan and Dan Roth. 2011.
\newblock \href {https://www.aclweb.org/anthology/P11-1056} {Exploiting
  syntactico-semantic structures for relation extraction}.
\newblock In \emph{Proceedings of the 49th Annual Meeting of the Association
  for Computational Linguistics: Human Language Technologies (ACL)}, pages
  551--560.

\bibitem[{Devlin et~al.(2019)Devlin, Chang, Lee, and
  Toutanova}]{devlin-etal-2019-bert}
Jacob Devlin, Ming-Wei Chang, Kenton Lee, and Kristina Toutanova. 2019.
\newblock \href {https://doi.org/10.18653/v1/N19-1423} {{BERT}: Pre-training of
  deep bidirectional transformers for language understanding}.
\newblock In \emph{Proceedings of the 2019 Conference of the North {A}merican
  Chapter of the Association for Computational Linguistics: Human Language
  Technologies, Volume 1 (Long and Short Papers) (NAACL)}, pages 4171--4186.

\bibitem[{Dixit and Al-Onaizan(2019)}]{dixit-al-onaizan-2019-span}
Kalpit Dixit and Yaser Al-Onaizan. 2019.
\newblock \href {https://doi.org/10.18653/v1/P19-1525} {Span-level model for
  relation extraction}.
\newblock In \emph{Proceedings of the 57th Annual Meeting of the Association
  for Computational Linguistics (ACL)}, pages 5308--5314.

\bibitem[{Dozat and Manning(2017)}]{dozat2017deep}
Timothy Dozat and Christopher~D. Manning. 2017.
\newblock \href {https://nlp.stanford.edu/pubs/dozat2017deep.pdf} {Deep
  biaffine attention for neural dependency parsing}.
\newblock In \emph{International Conference on Learning Representations
  (ICLR)}.

\bibitem[{Eberts and Ulges(2020)}]{spert}
Markus Eberts and Adrian Ulges. 2020.
\newblock \href {https://arxiv.org/abs/1909.07755} {Span-based joint entity and
  relation extraction with transformer pre-training}.
\newblock In \emph{24th European Conference on Artificial Intelligence (ECAI)}.

\bibitem[{Gupta et~al.(2016)Gupta, Sch{\"u}tze, and
  Andrassy}]{gupta-etal-2016-table}
Pankaj Gupta, Hinrich Sch{\"u}tze, and Bernt Andrassy. 2016.
\newblock \href {https://www.aclweb.org/anthology/C16-1239} {Table filling
  multi-task recurrent neural network for joint entity and relation
  extraction}.
\newblock In \emph{Proceedings of {COLING} 2016, the 26th International
  Conference on Computational Linguistics: Technical Papers (COLING)}, pages
  2537--2547.

\bibitem[{Li and Ji(2014)}]{li-ji-2014-incremental}
Qi~Li and Heng Ji. 2014.
\newblock \href {https://doi.org/10.3115/v1/P14-1038} {Incremental joint
  extraction of entity mentions and relations}.
\newblock In \emph{Proceedings of the 52nd Annual Meeting of the Association
  for Computational Linguistics (Volume 1: Long Papers) (ACL)}, pages 402--412.

\bibitem[{Li et~al.(2019)Li, Yin, Sun, Li, Yuan, Chai, Zhou, and
  Li}]{li-etal-2019-entity}
Xiaoya Li, Fan Yin, Zijun Sun, Xiayu Li, Arianna Yuan, Duo Chai, Mingxin Zhou,
  and Jiwei Li. 2019.
\newblock \href {https://doi.org/10.18653/v1/P19-1129} {Entity-relation
  extraction as multi-turn question answering}.
\newblock In \emph{Proceedings of the 57th Annual Meeting of the Association
  for Computational Linguistics (ACL)}, pages 1340--1350.

\bibitem[{Liu et~al.(2019)Liu, Meng, Zhang, Xu, Chen, and
  Zhou}]{liu-etal-2019-gcdt}
Yijin Liu, Fandong Meng, Jinchao Zhang, Jinan Xu, Yufeng Chen, and Jie Zhou.
  2019.
\newblock \href {https://doi.org/10.18653/v1/P19-1233} {{GCDT}: A global
  context enhanced deep transition architecture for sequence labeling}.
\newblock In \emph{Proceedings of the 57th Annual Meeting of the Association
  for Computational Linguistics (ACL)}, pages 2431--2441.

\bibitem[{Loshchilov and Hutter(2019)}]{loshchilov2018decoupled}
Ilya Loshchilov and Frank Hutter. 2019.
\newblock \href {https://openreview.net/forum?id=Bkg6RiCqY7} {Decoupled weight
  decay regularization}.
\newblock In \emph{International Conference on Learning Representations
  (ICLR)}.

\bibitem[{Luan et~al.(2019)Luan, Wadden, He, Shah, Ostendorf, and
  Hajishirzi}]{luan-etal-2019-general}
Yi~Luan, Dave Wadden, Luheng He, Amy Shah, Mari Ostendorf, and Hannaneh
  Hajishirzi. 2019.
\newblock \href {https://doi.org/10.18653/v1/N19-1308} {A general framework for
  information extraction using dynamic span graphs}.
\newblock In \emph{Proceedings of the 2019 Conference of the North {A}merican
  Chapter of the Association for Computational Linguistics: Human Language
  Technologies, Volume 1 (Long and Short Papers) (NAACL)}, pages 3036--3046.

\bibitem[{Miwa and Bansal(2016)}]{miwa-bansal-2016-end}
Makoto Miwa and Mohit Bansal. 2016.
\newblock \href {https://doi.org/10.18653/v1/P16-1105} {End-to-end relation
  extraction using {LSTM}s on sequences and tree structures}.
\newblock In \emph{Proceedings of the 54th Annual Meeting of the Association
  for Computational Linguistics (Volume 1: Long Papers) (ACL)}, pages
  1105--1116.

\bibitem[{Miwa and Sasaki(2014)}]{miwa-sasaki-2014-modeling}
Makoto Miwa and Yutaka Sasaki. 2014.
\newblock \href {https://doi.org/10.3115/v1/D14-1200} {Modeling joint entity
  and relation extraction with table representation}.
\newblock In \emph{Proceedings of the 2014 Conference on Empirical Methods in
  Natural Language Processing ({EMNLP})}, pages 1858--1869.

\bibitem[{Nadeau and Sekine(2007)}]{Nadeau:07}
David Nadeau and Satoshi Sekine. 2007.
\newblock \href
  {https://www.semanticscholar.org/paper/A-survey-of-named-entity-recognition-and-Nadeau-Sekine/4a554da55fd9ff76c99e25d2ce937b225dc1100c}
  {A survey of named entity recognition and classification}.
\newblock \emph{Linguisticae Investigationes}, 30(1):3--26.

\bibitem[{Nguyen and Verspoor(2019)}]{NguyenV_ECIR2019}
Dat~Quoc Nguyen and Karin Verspoor. 2019.
\newblock \href {https://datquocnguyen.github.io/resources/ECIR2019.pdf}
  {{End-to-end neural relation extraction using deep biaffine attention}}.
\newblock In \emph{Proceedings of the 41st European Conference on Information
  Retrieval (ECIR)}.

\bibitem[{Paszke et~al.(2019)Paszke, Gross, Massa, Lerer, Bradbury, Chanan,
  Killeen, Lin, Gimelshein, Antiga, Desmaison, Kopf, Yang, DeVito, Raison,
  Tejani, Chilamkurthy, Steiner, Fang, Bai, and Chintala}]{pytorch}
Adam Paszke, Sam Gross, Francisco Massa, Adam Lerer, James Bradbury, Gregory
  Chanan, Trevor Killeen, Zeming Lin, Natalia Gimelshein, Luca Antiga, Alban
  Desmaison, Andreas Kopf, Edward Yang, Zachary DeVito, Martin Raison, Alykhan
  Tejani, Sasank Chilamkurthy, Benoit Steiner, Lu~Fang, Junjie Bai, and Soumith
  Chintala. 2019.
\newblock \href
  {http://papers.neurips.cc/paper/9015-pytorch-an-imperative-style-high-performance-deep-learning-library.pdf}
  {Pytorch: An imperative style, high-performance deep learning library}.
\newblock In \emph{Advances in Neural Information Processing Systems 32
  (NIPS)}, pages 8024--8035. Curran Associates, Inc.

\bibitem[{Pradhan et~al.(2012)Pradhan, Moschitti, Xue, Uryupina, and
  Zhang}]{pradhan-etal-2012-conll}
Sameer Pradhan, Alessandro Moschitti, Nianwen Xue, Olga Uryupina, and Yuchen
  Zhang. 2012.
\newblock \href {https://www.aclweb.org/anthology/W12-4501} {Conll-2012 shared
  task: Modeling multilingual unrestricted coreference in ontonotes}.
\newblock In \emph{Joint Conference on EMNLP and CoNLL - Shared Task}, pages
  1--40.

\bibitem[{Ratinov and Roth(2009)}]{ratinov-roth-2009-design}
Lev Ratinov and Dan Roth. 2009.
\newblock \href {https://www.aclweb.org/anthology/W09-1119} {Design challenges
  and misconceptions in named entity recognition}.
\newblock In \emph{Proceedings of the Thirteenth Conference on Computational
  Natural Language Learning ({C}o{NLL}-2009)}, pages 147--155.

\bibitem[{Roth and Yih(2004)}]{roth-yih-2004-linear}
Dan Roth and Wen-tau Yih. 2004.
\newblock \href {https://www.aclweb.org/anthology/W04-2401} {A linear
  programming formulation for global inference in natural language tasks}.
\newblock In \emph{Proceedings of the Eighth Conference on Computational
  Natural Language Learning ({C}o{NLL}-2004) at {HLT}-{NAACL} 2004}, pages
  1--8.

\bibitem[{Strakov{\'a} et~al.(2019)Strakov{\'a}, Straka, and
  Hajic}]{strakova-etal-2019-neural}
Jana Strakov{\'a}, Milan Straka, and Jan Hajic. 2019.
\newblock \href {https://doi.org/10.18653/v1/P19-1527} {Neural architectures
  for nested {NER} through linearization}.
\newblock In \emph{Proceedings of the 57th Annual Meeting of the Association
  for Computational Linguistics (ACL)}, pages 5326--5331.

\bibitem[{Sun et~al.(2018)Sun, Wu, Lan, Sun, Wang, Lee, and
  Wu}]{sun-etal-2018-extracting}
Changzhi Sun, Yuanbin Wu, Man Lan, Shiliang Sun, Wenting Wang, Kuang-Chih Lee,
  and Kewen Wu. 2018.
\newblock \href {https://doi.org/10.18653/v1/D18-1249} {Extracting entities and
  relations with joint minimum risk training}.
\newblock In \emph{Proceedings of the 2018 Conference on Empirical Methods in
  Natural Language Processing (EMNLP)}, pages 2256--2265.

\bibitem[{Tjong Kim~Sang and
  De~Meulder(2003)}]{tjong-kim-sang-de-meulder-2003-introduction}
Erik~F. Tjong Kim~Sang and Fien De~Meulder. 2003.
\newblock \href {https://www.aclweb.org/anthology/W03-0419} {Introduction to
  the {C}o{NLL}-2003 shared task: Language-independent named entity
  recognition}.
\newblock In \emph{Proceedings of the Seventh Conference on Natural Language
  Learning at {HLT}-{NAACL} 2003}, pages 142--147.

\bibitem[{Vaswani et~al.(2017)Vaswani, Shazeer, Parmar, Uszkoreit, Jones,
  Gomez, Kaiser, and Polosukhin}]{vaswani2017attention}
Ashish Vaswani, Noam Shazeer, Niki Parmar, Jakob Uszkoreit, Llion Jones,
  Aidan~N Gomez, {\L}ukasz Kaiser, and Illia Polosukhin. 2017.
\newblock \href
  {https://papers.nips.cc/paper/7181-attention-is-all-you-need.pdf} {Attention
  is all you need}.
\newblock In \emph{Advances in neural information processing systems (NIPS)},
  pages 5998--6008.

\bibitem[{Wadden et~al.(2019)Wadden, Wennberg, Luan, and
  Hajishirzi}]{Wadden2019EntityRA}
David Wadden, Ulme Wennberg, Yi~Luan, and Hannaneh Hajishirzi. 2019.
\newblock \href {https://doi.org/10.18653/v1/D19-1585} {Entity, relation, and
  event extraction with contextualized span representations}.
\newblock In \emph{Proceedings of the 2019 Conference on Empirical Methods in
  Natural Language Processing and the 9th International Joint Conference on
  Natural Language Processing (EMNLP-IJCNLP)}, pages 5784--5789.

\bibitem[{Wolf et~al.(2019)Wolf, Debut, Sanh, Chaumond, Delangue, Moi, Cistac,
  Rault, Louf, Funtowicz, and Brew}]{Wolf2019HuggingFacesTS}
Thomas Wolf, Lysandre Debut, Victor Sanh, Julien Chaumond, Clement Delangue,
  Anthony Moi, Pierric Cistac, Tim Rault, R'emi Louf, Morgan Funtowicz, and
  Jamie Brew. 2019.
\newblock \href {https://arxiv.org/abs/1910.03771} {Huggingface's transformers:
  State-of-the-art natural language processing}.
\newblock \emph{ArXiv}, abs/1910.03771.

\bibitem[{Zelenko et~al.(2003)Zelenko, Aone, and Richardella}]{Zelenko:03}
Dmitry Zelenko, Chinatsu Aone, and Anthony Richardella. 2003.
\newblock \href {http://www.jmlr.org/papers/volume3/zelenko03a/zelenko03a.pdf}
  {Kernel methods for relation extraction}.
\newblock \emph{Journal of Machine Learning Research}, 3:1083–1106.

\bibitem[{Zhang et~al.(2017)Zhang, Zhang, and Fu}]{zhang-etal-2017-end}
Meishan Zhang, Yue Zhang, and Guohong Fu. 2017.
\newblock \href {https://doi.org/10.18653/v1/D17-1182} {End-to-end neural
  relation extraction with global optimization}.
\newblock In \emph{Proceedings of the 2017 Conference on Empirical Methods in
  Natural Language Processing (EMNLP)}, pages 1730--1740.

\bibitem[{Zhou et~al.(2005)Zhou, Su, Zhang, and
  Zhang}]{zhou-etal-2005-exploring}
GuoDong Zhou, Jian Su, Jie Zhang, and Min Zhang. 2005.
\newblock \href {https://doi.org/10.3115/1219840.1219893} {Exploring various
  knowledge in relation extraction}.
\newblock In \emph{Proceedings of the 43rd Annual Meeting of the Association
  for Computational Linguistics (ACL)}, pages 427--434.

\end{thebibliography}
\bibliographystyle{acl_natbib}

\newpage
\clearpage
\appendix

\section*{Appendix A: Major Hyper-Parameters}

This section contains a list of major hyper-parameters in our model, as shown in Table \ref{tab:hyper}. For both CoNLL04 and ACE05, we used the same hyperparameters with an exception to batch size, owing to the difference in the data scale. 
We applied dropout to $\bm{e}_{w_i}$ and $f(\bm{e}_{w_{{\rm first}(i-1)}}, \cdots, \bm{e}_{w_{i-1}})$ in Equation 2 and $\bm{z}_i$ in Equation 5.

\begin{table}[ht]
    \centering
    \begin{tabular}{lr}
      \Xhline{3\arrayrulewidth}    
         \multicolumn{1}{c}{Parameter} & \multicolumn{1}{c}{Value}  \\
         \hline
         \# dims of token embeddings $|\bm{e}_w|$ & 768 \\
         \# dims of label embeddings $|\bm{l}_y|$ & 50 \\
         \# dims of relation attention $d_{\rm att}$ & 20 \\
         \hline
         learning rate (BERT encoder) & $5 \times 10^{-5}$ \\
         learning rate (others) & $1 \times 10^{-3}$ \\
         dropout rate & $0.3$ \\
         warm-up period & $0.2$ \\
         total number of epochs & $30$ \\
      \Xhline{3\arrayrulewidth}         
    \end{tabular}
    \caption{Hyper-parameter settings. We adopted a learning rate scheduler that increased learning rates linearly from $0$ to $5 \times 10^{-5}$ in the warm-up period of $0.2 \times \mbox{(total number of epochs)}$, and then decreased learning rates using a cosine function.}
    \label{tab:hyper}
\end{table}

\section*{Appendix B: Error Inspection}
\label{sec:err}
\begin{table*}[t]
    \centering

    \begin{tabular}{ll}
    \Xhline{3\arrayrulewidth}
    \multicolumn{2}{c}{\textbf{(a)} Incorrect NE span} \\
    \hline
    \multirow{2}{*}{Ground truth} & Text of the statement issued by the \textcolor{red}{[}Organization of the Oppressed on Earth\textcolor{red}{]}\textsubscript{\sc Org}   \\
    &  claiming U. S. Marine Lt. William R. Higgins was hanged. \\
    \hdashline
    \multirow{2}{*}{Prediction} & Text of the statement issued by the \textcolor{red}{[}Organization of the Oppressed\textcolor{red}{]}\textsubscript{\sc Org} on Earth \\
    & claiming U. S. Marine Lt. William R. Higgins was hanged. \\
    \Xhline{2\arrayrulewidth}
    \multicolumn{2}{c}{\textbf{(b)} Incorrect NE type} \\
    \hline
    \multirow{3}{*}{Ground truth} & Manygate Management said Ogdon died peacefully after going into a coma \\
    & following his admission to London's \textcolor{red}{[}Charing Cross Hospital\textcolor{red}{]}$_\text{\sc Loc}$ Monday \\ & for bronchopneumonia. \\   
    \hdashline
    \multirow{3}{*}{Prediction} & Manygate Management said Ogdon died peacefully after going into a coma \\
    & following his admission to London's \textcolor{red}{[}Charing Cross Hospital\textcolor{red}{]}$_\text{\sc Org}$ Monday\\
    & for bronchopneumonia. \\   
    \Xhline{2\arrayrulewidth}
    \multicolumn{2}{c}{\textbf{(c)} Lack of knowledge} \\
    \hline
    \multirow{2}{*}{Sentence} & High winds blew on the east slopes of the \textcolor{red}{[}Rocky Mountains\textcolor{red}{]}$_\text{\sc Loc}$ in \textcolor{red}{[}Montana\textcolor{red}{]}$_\text{\sc Loc}$, \\ & with winds gusting to near 50 mph at \textcolor{red}{[}Livingston\textcolor{red}{]}$_\text{\sc Loc}$.\\
    \hdashline
    \multirow{3}{*}{Ground Truth} & $\mathrm{(\text{Rocky Mountains}_{\sc Loc}, LocatedIn, \text{Montana}_{\sc Loc})}$ \\
    & $\mathrm{(\text{Livingston}_{\sc Loc}, LocatedIn, \text{Montana}_{\sc Loc})}$ \\
    & $\mathrm{(\text{Livingston}_{\sc Loc}, LocatedIn, \text{Rocky Mountains}_{\sc Loc})}$\\
    \hdashline
    \multirow{2}{*}{Prediction} & $\mathrm{(\text{Rocky Mountains}_{\sc Loc}, LocatedIn, \text{Montana}_{\sc Loc})}$ \\
    & $\mathrm{(\text{Livingston}_{\sc Loc}, LocatedIn, \text{Montana}_{\sc Loc})}$ \\
    \Xhline{2\arrayrulewidth}
    \multicolumn{2}{c}{\textbf{(d)} Lack of global constraints} \\
    \hline
    \multirow{2}{*}{Sentence}  & \textcolor{red}{[}Soviet\textcolor{red}{]}$_\text{\sc Loc}$ Foreign \textcolor{red}{[}Eduard A. Shevardnadze\textcolor{red}{]}$_\text{\sc Peop}$ is to visit \textcolor{red}{[}China\textcolor{red}{]}$_\text{\sc Loc}$ next month \\
    & to pave the way for the first Chinese - Soviet summit in 30 years ... \\
    \hdashline
    Ground Truth & $\mathrm{(\text{Eduard A. Shevardnadze}_{Peop}, LiveIn, \text{Soviet}_{Loc})}$ \\   
    \hdashline
    \multirow{2}{*}{Prediction} & $\mathrm{(\text{Eduard A. Shevardnadze}_{Peop}, LiveIn, \text{Soviet}_{Loc})}$ \\  
    & $\mathrm{(\text{Eduard A. Shevardnadze}_{Peop}, LiveIn, \text{China}_{Loc})}$ \\  
    \Xhline{3\arrayrulewidth}      
    \end{tabular}
    \caption{Typical error cases of the proposed method on the CoNLL04 test set. Cases \textbf{(a)} and \textbf{(b)} are errors caused by the NER model; and Cases \textbf{(c)} and \textbf{(d)} are those caused by the RE model. Each RE case shows the sentence with the NE labels in the first line, followed by relation tuples of the ground truth annotation and the prediction.} 
    \label{tab:err}
\end{table*}

This section contains descriptions for incorrect predictions of the proposed method to delineate future directions for improvement. Table \ref{tab:err} summarizes typical errors of the proposed method found in the CoNLL04 test set.

\paragraph{Incorrect NE span} Cases where the NER model predicts slightly incorrect spans. Typical errors of this category involve adding/missing a nearby phrase of an NE span. Table \ref{tab:err} (a) is an example where the phrase ``on Earth'' can be interpreted as a prepositional phrase or a part of a proper noun.

\paragraph{Incorrect NE type} Cases where the NER model predicts incorrect NE labels for entity mentions. These cases usually occur when an entity can be interpreted with different NE types. Table \ref{tab:err} (b) illustrates that ``Charing Cross Hospital'' can be categorized as \textsc{Organization} if we look at the NE alone, but is actually annotated as \textsc{Location} in the context (indicating the location of the event `died').

\paragraph{Lack of Knowledge} Cases where the RE model fails to recognize implicit relations. In Table \ref{tab:err} (c), it is not so easy to recognize the relation instance $\mathrm{(\text{Livingston}_{Loc}, LocatedIn, \text{Montana}_{Loc})}$ only from the sentence without the knowledge about the entities `Livingston' and `Montana'. Fortunately, the RE model could predict the relation instance correctly in this example. However, it is even more difficult to infer the relation instance $\mathrm{(\text{Livingston}_{Loc}, LocatedIn, \text{Rocky Mountains}_{Loc})}$ from the text; we are not sure of the inclusion relation between ``Rocky Mountain'' and `Livingston' without the knowledge about the entities.

\paragraph{Lack of global constraints} Cases where the RE model could avoid incorrect RE instances with constraints. As shown in Table \ref{tab:err} (d), the model infers that the same person lives in two different places (Soviet and China). the proposed method cannot consider associations between relation predictions explicitly because relation labels are predicted independently of each other.

\section*{Appendix C: Predicted Examples for Multi-Sentence NER }
\label{doclevel_example}

This section contains several typical predicted examples showing the effectiveness of the multi-sentence NER model, as shown in Table \ref{tab:coref_examples}.
\begin{table*}[t]
    \centering
    \begin{tabular}{lll}
    \Xhline{3\arrayrulewidth} 
    \multicolumn{3}{c}{\textbf{CoNLL03}} \\
    \Xhline{2\arrayrulewidth}
    \multicolumn{3}{c}{\texttt{Location} (single), \texttt{Organization} (multi), \texttt{Organization} (gold)} \\
\multicolumn{3}{l}{\textit{Charleroi} ( Belgium ) 75 Estudiantes Madrid ( Spain ) 82 ( 34 - 35 )}\\
 \multicolumn{3}{l}{Leading scorers: \textbf{Charleroi} - Eric Cleymans 18, Ron Ellis 18, Jacques Stas 14 } \\
    \Xhline{2\arrayrulewidth}
    \multicolumn{3}{c}{\texttt{Person} (single), \texttt{Organization} (multi), \texttt{Organization} (gold)} \\
    \multicolumn{3}{l}{\textbf{Tambang Timah} at \$ 15. 575 in London.}\\
    \multicolumn{3}{l}{LONDON 1996 - 12 - 07}\\
 \multicolumn{3}{l}{
PT \textit{Tambang Timah} closed at \$ 15. 575 per GDR in London on Friday. } \\
    \Xhline{3\arrayrulewidth} 
    \multicolumn{3}{c}{\textbf{ACE05}} \\
    \Xhline{2\arrayrulewidth}
    \multicolumn{3}{c}{\texttt{Person} (single), \texttt{Organization} (multi), \texttt{Organization} (gold)} \\
 \multicolumn{3}{l}{\textit{North Korea} has told American lawmakers it already has nuclear weapons ... } \\
    \multicolumn{3}{l}{``\textbf{They} admitted to having just about completed the reprocessing of 8, 000 rods,'' said ...} \\
    \Xhline{2\arrayrulewidth}
    \multicolumn{3}{c}{\texttt{Geographical Entity} (single), \texttt{Person} (multi), \texttt{Person} (gold)} \\
 \multicolumn{3}{l}{...today's \textit{Southern voters} are ``children of Democrats who are not swayed by the same things ... } \\
    \multicolumn{3}{l}{\textbf{They} certainly are susceptible to the Republican message. ''} \\
    \Xhline{3\arrayrulewidth}
    \end{tabular}
    \caption{Typical examples where the proposed NER model showed improvements with multi-sentence input. A \textbf{boldface} word presents a target for predicting the NE label, and an \textit{italic} word is a coreference to the boldface word. NE labels shown on top of each example are the prediction from single-sentence input; that from multi-sentence input; and the ground-truth label. We can infer that the model with multi-sentence input made correct predictions, looking at coreferential words.}
    \label{tab:coref_examples}
\end{table*}

\end{document}